\documentclass[journal, twoside, web]{ieeecolor}
\usepackage{generic}
\usepackage{graphicx}
\usepackage{graphics} 
\usepackage{epsfig} 
\usepackage{times} 
\usepackage{amsmath} 
\usepackage{amssymb}  
\usepackage{tikz}
\usepackage{standalone}
\usepackage{bondgraphs}
\usepackage{url}
\usepackage{blox}
\usepackage{gensymb}
\usepackage{booktabs,amsfonts,dcolumn}
\usepackage{blindtext}
\usepackage{rotating}
\usetikzlibrary{shapes,arrows}
\usetikzlibrary{calc, patterns, decorations.pathmorphing, decorations.markings, shapes.geometric, arrows, positioning, angles}

\mathchardef\mhyphen="2D
\mathchardef\mslash="202F
\def\tablename{Table}
\usepackage{textcomp}
\usepackage{hyperref}
\usepackage{lipsum}
\allowdisplaybreaks

\usepackage{multirow}
\usepackage{accents}
\newcommand{\ubar}[1]{\underaccent{\bar}{#1}}
\newcolumntype{L}[1]{>{\raggedright\arraybackslash}p{#1}}
\newcolumntype{C}[1]{>{\centering\arraybackslash}p{#1}}
\newcolumntype{R}[1]{>{\raggedleft\arraybackslash}p{#1}}

\definecolor{b1}{RGB}{0, 104, 179}
\definecolor{r1}{RGB}{148, 44, 44}

\title{A Complex Stiffness Human Impedance Model with Customizable Exoskeleton Control}
\author{Binghan He, Huang Huang, Gray C. Thomas,~\IEEEmembership{Member,~IEEE}, and Luis Sentis,~\IEEEmembership{Member,~IEEE} 
\thanks{This work was supported by the U.S. Government and NASA Space Technology Research Fellowship NNX15AQ33H. 
The authors thank the members of the Human Centered Robotics Lab, The University of Texas at Austin, and Apptronik Systems for their support. 
\emph{(Corresponding author: Binghan He.)}}
\thanks{Binghan He and Huang Huang are with the Department of Mechanical Engineering, The University of Texas at Austin, Austin, TX 78712 USA (e-mail: binghan@utexas.edu; huangh@utexas.edu)}
\thanks{Gray C. Thomas is with the Department of Electrical and Computer Engineering, University of Michigan, Ann Arbor, MI 48109 USA (e-mail: gcthomas@umich.edu)}
\thanks{Luis Sentis is with the Department of Aerospace Engineering and Engineering Mechanics, The University of Texas at Austin, Austin, TX 78712 USA (e-mail: lsentis@austin.utexas.edu)}
}

\newcommand\copyrighttext{%
  \scriptsize\rm
  Accepted for publication in Transactions on Neural Systems and Rehabilitation Engineering (TNSRE)
  \textcopyright 2020 IEEE. Personal use of this material is permitted. Permission from IEEE must be obtained for all other uses, in any current or future media, including reprinting/republishing this material for advertising or promotional purposes, creating new collective works, for resale or redistribution to servers or lists, or reuse of any copyrighted component of this work in other works.
  DOI: 10.1109/TNSRE.2020.3027501
}
\newcommand\copyrightnotice{%
\begin{tikzpicture}[remember picture,overlay]
\node[anchor=south,yshift=15pt] at (current page.south)
{\fbox{\parbox{\dimexpr\textwidth-\fboxsep-\fboxrule\relax}{\copyrighttext}}};
\end{tikzpicture}%
}
\pubid{}

\begin{document}

\maketitle
\copyrightnotice

\vspace{-10pt}
\begin{abstract}
The natural impedance, or dynamic relationship between force and motion, of a human operator can determine the stability of exoskeletons that use interaction-torque feedback to amplify human strength.
While human impedance is typically modelled as a linear system, our experiments on a single-joint exoskeleton testbed involving 10 human subjects show evidence of nonlinear behavior: a low-frequency asymptotic phase for the dynamic stiffness of the human that is different than the expected zero, and an unexpectedly consistent damping ratio as the stiffness and inertia vary.
To explain these observations, this paper considers a new frequency-domain model of the human joint dynamics featuring complex value stiffness comprising a real stiffness term and a hysteretic damping term.
Using a statistical F-test we show that the hysteretic damping term is not only significant but is even more significant than the linear damping term.
Further analysis reveals a linear trend linking hysteretic damping and the real part of the stiffness, which allows us to simplify the complex stiffness model down to a 1-parameter system.
Then, we introduce and demonstrate a customizable fractional-order controller that exploits this hysteretic damping behavior to improve strength amplification bandwidth while maintaining stability, and explore a tuning approach which ensures that this stability property is robust to muscle co-contraction for each individual. 
\end{abstract} 


\section{Introduction}
\IEEEPARstart{W}{hile} the concept of a personal augmentation device or exoskeleton has a long history \cite{YagnNicolas1890PatentUSpatent, MakinsonBodineFitck1969report,KazerooniGuo1993JDSMC}, a system which delivers on the dream of transparent interaction, of ``feeling like the system is not there,'' through amplification of sensed human interaction forces is still an ambitious goal of force control technology today \cite{Kazerooni2005IROS,DollarHerr2008TRO, JacobsenOlivier2014Patent, FontanaVertechyMarcheschiSalsedoBergamasco2014RAM}. Unlike other assistive exoskeletons that help perform predictable behaviors \cite{ZhangFiersWitteJacksonPoggenseeAtkesonCollins2017Science, LeeKimBakerLongKaravasMenardGalianaWalsh2018JNR}, provide rehabilitation therapy \cite{KongMoonJeonTomizuka2010TMech, KimDeshpande2017IJRR}, or adjust natural dynamics in a helpful way \cite{LvGregg2018TCST}, human amplification exoskeletons \cite{Kazerooni2005IROS, LeeLeeKimHanShinHan2014Mechatronics} assist users, whether patients or healthy people, by amplifying their strength (and power) through feedback control. But this type of feedback control introduces a risk of instability (a risk that was first noted in the field of impedance control for physical human--robot interactions \cite{BuergerHogan2007TRO}). Since the exoskeleton is in a feedback interconnection with the human, a model of the human's dynamic behavior plays a critical role in determining the stability of the closed loop amplification exoskeleton system \cite{HeThomasPaineSentis2019ACC, HuangCappelThomasHeSentis2020ACC}.

Time and gait-phase varying models of the human joint impedance \cite{RouseHargrovePerreaultKuiken2014TNSRE, LeeHogan2015TNSRE, LeeRouseKrebs2016JTEHM, ShorterRouse2018TNSRE, ShorterRouse2019TBME} have been pursued by the bio-mechanics and wearable robotics communities to better replicate human lower-limb behavior. 
Among all different kinds of impedance model of an individual human joint, perhaps the most popular one is the mass-spring-damper model---with the additional non-linearity that the spring stiffness of the human joint can be modified by both voluntary muscle contractions or external torques exerted on the joint \cite{BennettHollerbachXuHunter1992EBR}. 
Several studies demonstrated a linear relationship between the stiffness 
(not to be confused with quasi-stiffness \cite{RouseGreggHargroveSensinger2012TBE})
of the human (found by fitting a linear mass-spring-damper model for a single joint) and an external torque \cite{AgarwalGottlieb1977JBE, GottliebAgarwal1978JB, ZahalakHeyman1979JBE, CannonZahalak1982JB, HunterKearney1982JB}. 
Joint damping has also been shown to increase with muscle contractions \cite{BeckerMote1990JBE} and external torques \cite{WeissHunterKearney1988JB}. 
A linear relationship between damping and external torque has also been reported for the same human joints, but it is statistically weaker than the strong linear relationship between stiffness and external torques \cite{AgarwalGottlieb1977JBE, HunterKearney1982JB}. 
However, it is not clear from the literature that a linear relationship between damping and stiffness in human joints does exist. 

\pubidadjcol

Yet inconsistencies in the variable mass-spring-damper model remain  \cite{SobhanitehraniJalaleddiniKearney2017TNSRE}, and the empirical observation that a relatively consistent damping ratio is maintained in some joints (notably the human elbow \cite{HeThomasPaineSentis2019ACC} and arm \cite{PerreaultKirschCrago2004EBR}) even as joint stiffness and inertia vary is one such anomaly. Frequency domain identification of the ankle joint impedance \cite{AgarwalGottlieb1977JBE, GottliebAgarwal1978JB} also shows a consistent damping ratio within the range from $0.22$ to $0.49$. This damping ratio consistency on the ankle is also supported by the fact that the ankle damping ratio does not have significant change with large variations of mean external torques exerted on the subjects \cite{WeissHunterKearney1988JB}. For upper limbs, a multi-joint impedance study on human arms \cite{PerreaultKirschCrago2004EBR} showed that the damping ratio of the minimally damped mode for the 2-D endpoint impedance in the transverse plane is distributed with a mean of $0.26$ and a standard deviation of $0.08$. Although this could be explained as humans adapting their damping to stabilize movement \cite{MilnerCloutier1993EBR}, a more detailed explanation of how humans achieve this consistency remains unclear.

One potential solution is to use hysteretic damping models. Hysteretic damping can correctly explain the behaviors shown in \cite{AgarwalGottlieb1977JBE, GottliebAgarwal1978JB, ZahalakHeyman1979JBE, CannonZahalak1982JB}, where the phase plots of the human stiffness have non-zero values at low frequencies. In particular, Ref. \cite[Fig.~6]{CannonZahalak1982JB} shows that the human elbow dynamic stiffness has a phase shift around $25 \degree$ for a wide range of low frequencies, thus contradicting the viscous damping hypothesis.
This type of phase behavior is explained in the field of structural mechanics by defining a hysteretic damping term whose damping coefficient is proportional to the inverse of the frequency \cite{BishopJohnson2011book}.

This paper introduces and validates a complex stiffness model for human elbow joint dynamics. The primary model validation experiment uses statistical F-tests to compare three dynamic stiffness models: a linear mass, spring, and viscous damper model, a nonlinear complex-stiffness-spring and mass model (that is, a spring, mass, and hysteretic damper model), and a combined model with mass, spring, and both viscous and hysteretic damping. This hysteretic damping explains the consistent damping ratio of the human--exoskeleton resonant peak even as the stiffness and exoskeleton inertia change---which is not well explained by the linear model. And it also explains the low frequency phase lag in human stiffness (previously observed in \cite{AgarwalGottlieb1977JBE, GottliebAgarwal1978JB, ZahalakHeyman1979JBE, CannonZahalak1982JB}). 

Using this new model, this paper introduces a customizable fractional-order controller designed to take full advantage of the low-frequency phase lag for each individual. Based on results from the previous test, a customized fractional order is chosen for each of three subjects such that the behavior is nearly oscillatory (marginally stable). The subjects then change their co-contraction level to illustrate the phenomenon of co-contraction induced instability and {subject dependent co-contraction relationships with stability}. The three subjects span the range of observed co-contraction--stability relationships. 

This paper builds significantly on our earlier conference presentation \cite{HeHuangThomasSentis2019IROS}.
First, this study investigates the applicability of the model in the more general population ($\mathrm{N}=10$), whereas \cite{HeHuangThomasSentis2019IROS} only supported the model in one single individual ($\mathrm{N}=1$).
Second, this study proposes a novel power law relationship that is necessary to describe the range of behaviors observed in the wider population, whereas \cite{HeHuangThomasSentis2019IROS} related stiffness and hysteretic damping with a naive linear model.
Third, this study highlights the novel finding that there are fundamental differences in the stiffness--hysteretic damping relationships between subjects, and the critical influence this has on the problem of designing tests to pre-certify safety in exoskeleton controllers, whereas \cite{HeHuangThomasSentis2019IROS} had no inter-subject data. Finally, this study employs a physical implementation, tests the three most extreme subjects, and provides the first empirical validation that the combination of fractional order control and tuning to the hysteretic damping model can improve dynamical amplification (at $10$ $\mathrm{rad/s}$) by $81 \sim 88 \%$, whereas \cite{HeHuangThomasSentis2019IROS} only proposed the concept of using fractional-order controllers to exploit the hysteretic damping characteristics without any experimentation. 

The experimental protocol for model identification and  controller implementation was approved by the Institutional Review Board (IRB) at the University of Texas at Austin under study $\mathrm{No.} \ 2017 \mhyphen 10 \mhyphen 0006$. The informed consent forms were signed by all subjects.

\section{Modeling Methods}

\subsection{Apparatus}

\begin{figure} [!tbp]
\centering
\footnotesize
\def\svgwidth{.49\textwidth}
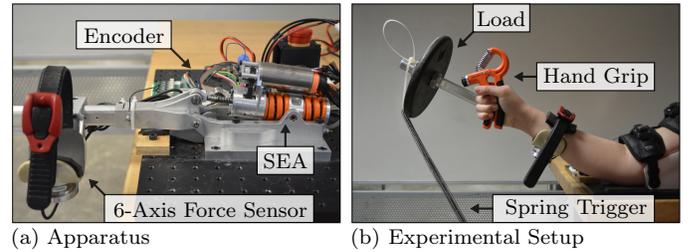
\caption{Experimental apparatus: the series elastic P0 exoskeleton featuring an ATI Mini40 force sensitive cuff and a P170 Orion air cooled series elastic actuator module acting through a simple 3 bar linkage. During all experiments, subjects apply forces to a adjustable hand grip to regulate their elbow stiffness. A spring trigger is only used for perturbation during the loop shaping experiments in Sec.~\ref{trigger}.}
\label{setup}
\end{figure}

For this study we employed the P0 series elastic elbow-joint exoskeleton (Apptronik Systems,\ Inc.,\ Austin,\ TX), as shown in Fig.~\ref{setup}. This exoskeleton has a moment of inertia of 0.1 $\rm{kg \cdot m^2}$ with no load on it but allows for attaching additional weights to it. A load, attached 0.45 m from the exoskeleton joint, is pictured in Fig.~\ref{setup}.(b). The contact force $f_c$ between the human and the exoskeleton is measured by a six-axis force/torque sensor situated below the white 3D printed ``cuff'' (which includes the adjustable strap that clamps the forearm). This force torque signal is cast as a torque ($\tau_c$) using the motion Jacobian $J$ of the sensor frame ($\tau_c = J^T f_c$). Rubber pads are adhered to the inside surfaces of the cuff and the cuff strap to improve user comfort. Joint position $\theta_e$ is directly measured by a dedicated encoder at the exoskeleton joint. The series elastic actuator (SEA) has a spring force control bandwidth of $10$ $\mathrm{Hz}$ and provides high fidelity actuator torque $\tau_s$ tracking using the disturbance observer of \cite{PaineOhSentis2014TMech}. 

\begin{figure}[!tbp]
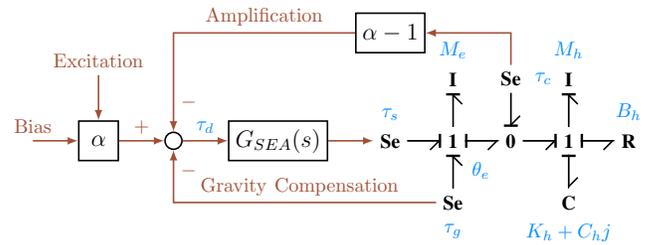

\centering
\includestandalone[width=.50\textwidth]{block-diagram}
\caption{Block diagram consisting of amplification, gravity compensation and experimental perturbation. The dynamics of the human with the exoskeleton are expressed as a bond graph with the effort sources $\tau_s$, $\tau_c$ and $\tau_g$.}
\label{block-diagram}
\end{figure}

In parallel with an excitation chirp command (which essentially performs system identification of the human subject), a gravity compensation controller, a human amplification controller, and a bias torque comprise the desired actuator torque signal. The gravity compensation controller takes the measurement of $\theta_e$ to calculate and compensate the gravity torque $\tau_g$ acting on the exoskeleton system. The human amplification controller takes the measurement of $\tau_c$ and multiplies it by a term equal to negative $\alpha-1$, where $\alpha \geq 1$ is an amplification factor. With the assistance of actuator torques produced from the amplification command, the human contact forces with the exoskeleton are amplified by the factor $\alpha$. 

\subsection{Models}

We use three models describing human-exoskeleton interactions in our statistical tests. As preliminaries, we first define $K_{h}$ as the (real-valued) human elbow-joint apparent stiffness, $H_{h}$ as the human elbow-joint hysteretic damping, $B_{h}$ as the human elbow-joint viscous damping, $M_{h}$ as the moment of inertia of the human, and $M_{e}$ as the moment of inertia of the exoskeleton. With the amplification control specified by the factor $\alpha$, the subject feels an attenuated inertia $M_{e} / \alpha$ from the interaction with the exoskeleton. Therefore, we also define the perceived inertia $M_{h \mhyphen e \mslash \alpha} \triangleq
M_{h} + M_{e} / \alpha$ at the elbow joint. 

The first model is a passive linear model with viscous damping and stiffness:
\begin{equation}
S_{h \mhyphen e \mslash \alpha} (s) = M_{h \mhyphen e \mslash \alpha} s^2 + B_{h} s + K_{h}.  \label{M1} \tag{M1}
\end{equation}
Replacing the viscous damping in \eqref{M1} by a hysteretic damping we arrive at our second model:
\begin{equation}
S_{h \mhyphen e \mslash \alpha} (s) = M_{h \mhyphen e \mslash \alpha} s^2 + H_{h} j + K_{h},  \label{M2} \tag{M2} 
\end{equation}
where a complex stiffness appears. Finally, to generalize \eqref{M1} and \eqref{M2}, we consider a third model with both viscous and hysteretic damping:
\begin{equation} 
S_{h \mhyphen e \mslash \alpha} (s) = M_{h \mhyphen e \mslash \alpha} s^2 + B_{h} s + H_{h} j + K_{h}.  \label{M3} \tag{M3}
\end{equation}

In order to take advantage of the clean human cuff sensor signal, we express these models in terms of the dynamic stiffness of the human alone, $S_h(s) = \tau_c(s)/\theta_e(s)$, using the following three equalities to learn the model parameters of \eqref{M1}--\eqref{M3} respectively:
\begin{align}
& S_{h} (s) = M_{h} s^2 + B_{h} s + K_{h},  \label{Sh1}  \\ 
& S_{h} (s) = M_{h} s^2 + H_{h} j + K_{h},  \label{Sh2}  \\
& S_{h} (s) = M_{h} s^2 + B_{h} s + H_{h} j + K_{h}.  \label{Sh3}  
\end{align}
The original transfer function can be recovered by adding in the exoskeleton inertia term
$
    S_{h \mhyphen e \mslash \alpha} (s) = S_h(s)+\frac{1}{\alpha}M_{e}s^2.
$
By re-casting the parameter estimation problem as the problem of estimating this re-creation of $S_{h \mhyphen e \mslash \alpha}(s)$, we can take advantage of the clean sensor data and avoid various corrupting effects in the $\tau_s(s)$ signal. Since the actual exoskeleton's dynamics are bypassed, the potential influence of unmodeled exoskeleton damping on the estimated parameters is eliminated.

\subsection{Experimental Protocol for Modeling}
The modeling study consists of nine perturbation experiments with $10$ healthy subjects between the ages of $21$-$29$, where subjects A-E are females and subjects F-J are males. 

The nine experiments are separated into three groups of three experiments. The three experiments in each group are conducted with a $4.5$ $\mathrm{kg}$ load and an $\alpha$ value of $1$ (corresponding to no amplification), $2$, and $4$. The gravity torque of the load and the exoskeleton itself are cancelled out by the gravity compensation feature of the controller, while the total inertia is attenuated by a factor of $\alpha$ due to the amplification feature.

Each of the three experimental groups are differently perturbed to achieve variation in elbow stiffnesses. Because the stiffness is determined by both muscle co-contraction and contraction to resist an external torque, we induce variation in stiffness by having each subject squeeze an adjustable force hand grip and by applying a bias torque from the actuator. The three experimental groups are divided into pairs of gripping forces and bias torques. The first group uses a $10$ $\mathrm{kg}$ gripping force and a $0$ $\mathrm{Nm}$ bias torque. The second group uses $14$ $\mathrm{kg}$ and $4$ $\mathrm{Nm}$. And the third group uses $27$ $\mathrm{kg}$ and $8$ $\mathrm{Nm}$. 

\setlength{\tabcolsep}{0.0pt}
\begin{table} [!tbp]
\caption{Modeling Experiment Parameters}
\centering
\scriptsize
\begin{tabular}{L{0.6cm} C{1.1cm} C{1.1cm} C{1.1cm} C{1.1cm} C{1.4cm} C{2.4cm}} \toprule
 \multirow{2}{*}{$\mathrm{Exp}$} & \multirow{2}{*}{$\alpha$} & $\mathrm{Load}$ & $\mathrm{Grip}$ & $\mathrm{Bias}$ & $\mathrm{Amplitude}$ &  $\mathrm{Frequency \ Range}$ \\ 
                      &                           & $\mathrm{(kg)}$ & $\mathrm{(kg)}$ & $\mathrm{(Nm)}$ &      $\mathrm{(Nm)}$ &    $\mathrm{(rad/s\  to\ rad/s)}$ \\ [.5ex] 
 \midrule
 $1$ & $1$ & \multirow{3}{*}{$4.5$} & \multirow{3}{*}{$10$} & \multirow{3}{*}{$0$} & \multirow{3}{*}{$2$} & \multirow{3}{*}{\footnotesize$2$\tiny$\times 10 ^ {0}$\footnotesize~$\mathrm{to}$~$2$\tiny$\times 10 ^ {0.9}$} \\
 $2$ & $2$ &   &   &   &   &   \\
 $3$ & $4$ &   &   &   &   &   \\ \midrule
 $4$ & $1$ & \multirow{3}{*}{$4.5$} & \multirow{3}{*}{$14$} & \multirow{3}{*}{$4$} & \multirow{3}{*}{$2$} & \multirow{3}{*}{\footnotesize$3$\tiny$\times 10 ^ {0}$\footnotesize~$\mathrm{to}$~$3$\tiny$\times 10 ^ {0.9}$} \\
 $5$ & $2$ &   &   &   &   & \\
 $6$ & $4$ &   &   &   &   & \\ \midrule
 $7$ & $1$ & \multirow{3}{*}{$4.5$} & \multirow{3}{*}{$27$} & \multirow{3}{*}{$8$} & \multirow{3}{*}{$2$} & \multirow{3}{*}{\footnotesize$4$\tiny$\times 10 ^ {0}$\footnotesize~$\mathrm{to}$~$4$\tiny$\times 10 ^ {0.9}$} \\
 $8$ & $2$ &   &   &   &   & \\
 $9$ & $4$ &   &   &   &   & \\ [0ex]
\bottomrule
\end{tabular} \label{exp-group}
\end{table}

Each of the nine perturbation experiments includes ten $60$-$\mathrm{sec}$ periods. The bias torque is gradually added during the first $5$ $\mathrm{sec}$ of each period while the subject raises the forearm to around a $45 \degree$ angle from the resting position and starts to squeeze the hand grip. Then, a sinusoidal perturbation signal is added for the next $10$ $\mathrm{sec}$. After the sinusoidal perturbation signal finishes, the bias torque is gradually subtracted for another $5$ $\mathrm{sec}$ with the subject bringing the arm back to the resting position and relaxing the hand. To avoid fatigue, the subject rests for the next $40$ $\mathrm{sec}$ in each period.
 
In order to capture the natural frequency of the human elbow joint wearing the exoskeleton, we set different values of the perturbation frequency for the different groups of experiments previously described. The three experimental groups use $2$ $\mathrm{rad / s}$, $3$ $\mathrm{rad / s}$, and $4$ $\mathrm{rad / s}$ for perturbation in the first time period. For other time periods, we set the perturbation frequencies to be $10 ^ {0.1}$ times the frequency of the previous perturbation. 

\begin{figure*}[!tbp]
    \centering
    \tiny
        \def\svgwidth{1.\textwidth}
    	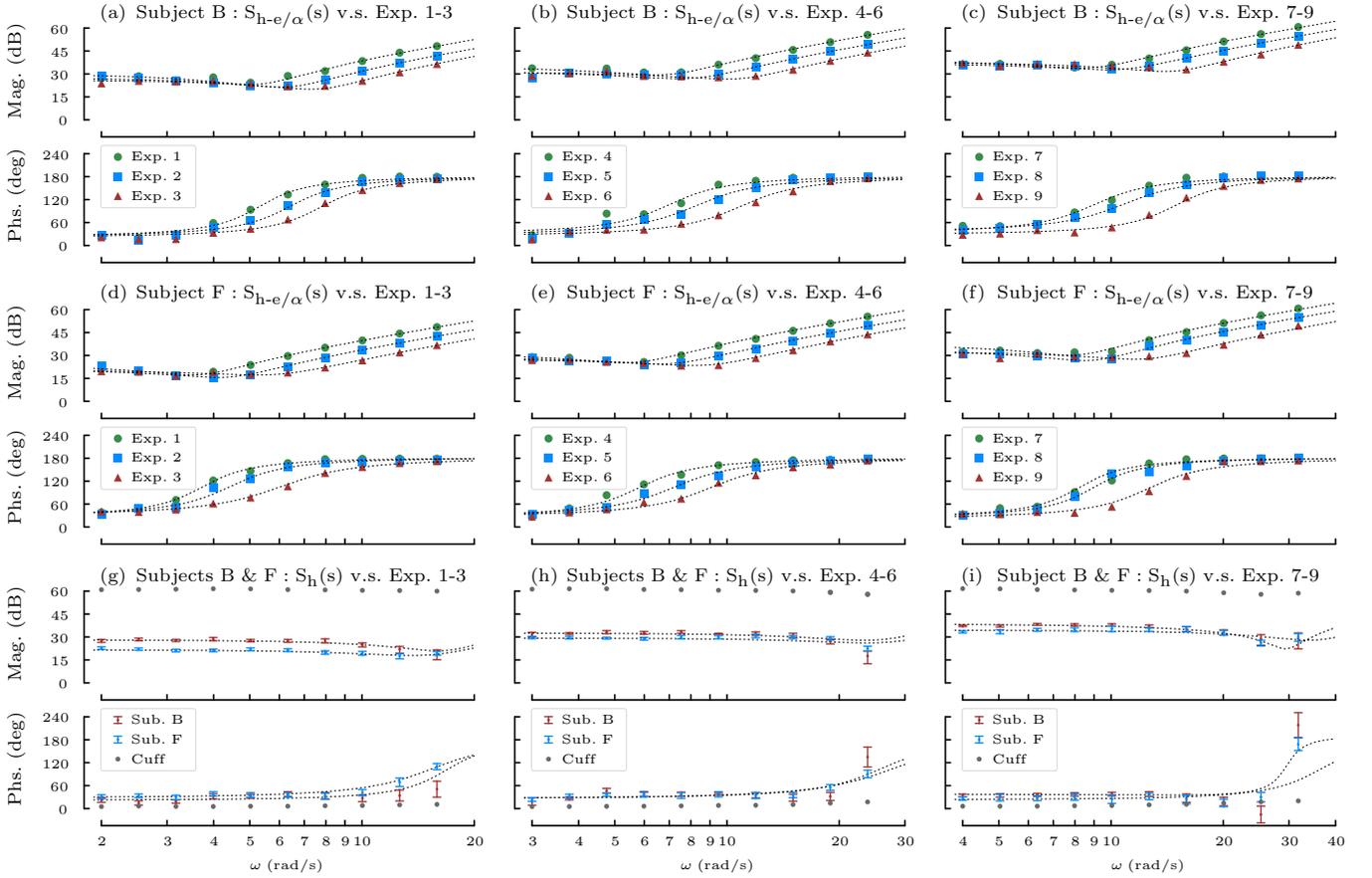
    \caption{Bode plots for $S_{h \mhyphen e \mslash \alpha} (s)$ showing all experiments for subject B in (a)-(c) and for subject F in (d)-(f). Bode plots for $S_{h} (s)$ in (g)-(i) showing the mean and standard error across each experimental group for subjects B and F. The gray dots in (g)-(i) show the dynamic stiffness of the cuff identified through a superposition test. The dash lines show the fitted curves using M3 and \eqref{Sh3}.}
    \label{fig:bode}
\end{figure*}

The amplitude of the sinusoidal perturbation signal is set to $2$ $\mathrm{Nm}$. However, after the perturbation frequency is higher than the natural frequency of $S_{h \mhyphen e \mslash \alpha} (s)$, the inertia effect $M_{h \mhyphen e \mslash \alpha}$ starts dominating the dynamic response and therefore the angle of displacement $\theta_e$ becomes less and less sensitive to the torque excitation. Thus, starting at the $\mathrm{8th}$ perturbation time period for each experiment, we increase the amplitude of the perturbation signal by $10 ^ {0.2}$ times in order to increase the sensitivity to the torque excitation. 

In the end, we identify the three models of $S_{h \mhyphen e \mslash \alpha}$ for all $90$ subject experiments using linear regression in the frequency domain obtained from time domain data. The parameters of all nine experimental settings are summarized in Tab.~\ref{exp-group}.

\subsection{Statistical Analysis}


Since we split each experiment into ten $60$-$\mathrm{sec}$ periods with $40$-$\mathrm{sec}$ resting time within each period, the response of $S_{h \mhyphen e \mslash \alpha} (s)$ to the sinusoidal perturbation in each period completely dies out before the next period. Therefore, for the purposes of statistical testing, we can safely assume statistical independence between any two single-frequency data points in each experiment. 

Regarding the $10$-$\mathrm{sec}$ sinusoidal perturbation within each period, only the data from the second $5$-$\mathrm{sec}$ part of the perturbation is used for calculating each frequency domain sample. Because the first $5$-$\mathrm{sec}$ perturbation time is greater than the $2\%$ settling time for all $S_{h \mhyphen e \mslash \alpha} (s)$ identified in our experiments, the output response reaches sinusoidal steady-state before entering the second $5$-$\mathrm{sec}$ perturbation time period.


For each experiment on each subject, we calculate the residual square sum ($\mathrm{RSS}$) for all three models, denoted as $\mathrm{R^{sub \mhyphen exp}_{M1}}$, $\mathrm{R^{sub \mhyphen exp}_{M2}}$ and $\mathrm{R^{sub \mhyphen exp}_{M3}}$ respectively, where $\mathrm{sub = A, \, B, \, \cdots, \, J}$ and $\mathrm{exp = 1, \, 2, \, \cdots, \, 9}$ are the indices of subjects and experiments. 
For $\mathrm{i = 1, \, 2, \, 3}$, let us define
\begin{align} \label{rss}
& \mathrm{R^{sub}_{Mi}} \triangleq \mathrm{\sum_{exp=1}^{9} R^{sub \mhyphen exp}_{Mi}},  \quad 
  \mathrm{R^{exp}_{Mi}} \triangleq \mathrm{\sum_{sub=A}^{J} R^{sub \mhyphen exp}_{Mi}},  \\
& \mathrm{R^{all}_{Mi}} \triangleq \mathrm{\sum_{exp=1}^{9} \sum_{sub=A}^{J} R^{sub \mhyphen exp}_{Mi}}.
\end{align}


In order to compare the significance of $B_{h} s$ and $H_{h} j$ in the human-exoskeleton interaction model, we conduct F-tests for each of the two three-parameter models (M1 and M2) against the generalizing four-parameter model (M3). 
Our F-statistic accounts for frequency domain data. For $\mathrm{i = 1, \, 2}$,
\begin{align} \label{ft}
\mathrm{F^{sub}_{Mi \mhyphen M3}} & = \mathrm{\frac{R^{sub}_{Mi} - R^{sub}_{M3}}{R^{sub}_{M3}} \cdot \frac{(2n-4) \cdot n_{exp}}{ \hfill (4-3) \cdot n_{exp}}},   \\ 
\mathrm{F^{exp}_{Mi \mhyphen M3}} & = \mathrm{\frac{R^{exp}_{Mi} - R^{exp}_{M3}}{R^{exp}_{M3}} \cdot \frac{(2n-4) \cdot n_{sub}}{ \hfill (4-3) \cdot n_{sub}}},   \\
\mathrm{F^{all}_{Mi \mhyphen M3}} & = \mathrm{\frac{R^{all}_{Mi} - R^{all}_{M3}}{R^{all}_{M3}} \cdot \frac{(2n-4) \cdot n_{exp} \cdot n_{sub}}{ \hfill (4-3) \cdot n_{exp} \cdot n_{sub}}}, 
\end{align}
where $\mathrm{n_{sub}}  = 10$ is the number of subjects, $\mathrm{n_{exp}}  = 9$ is the number of experiments per subject, $\mathrm{n} = 10$ is the number of complex value samples in the frequency domain, and the factor of two represents statistical independence between the real and imaginary parts of each sample. 

\section{Modeling Results} \label{sec:exp1_res}

\subsection{Frequency Domain Results} \label{phase}

The frequency data for the two most representative subjects are shown in Fig.~\ref{fig:bode}. Fig.~\ref{fig:bode}.(a)-(f) shows the Bode plots of $S_{h \mhyphen e \mslash \alpha} (s)$. Similarly to \cite[Fig.~4]{AgarwalGottlieb1977JBE}, \cite[Fig.~3]{GottliebAgarwal1978JB}, \cite[Fig.~2]{ZahalakHeyman1979JBE}, the phase for each experiment shows a non-zero value (near $30 \degree$ for subjects B and F) at low frequencies. This type of phase shift is very different from the phase shift values usually described by linear systems with viscous damping where the phase shift approaches zero as $\omega\rightarrow0$.

Since $S_h (s)$ is unaffected by changes of $M_{h \mhyphen e \mslash \alpha}$, we compute the statistics for all three experiments in each experimental group. Fig.~\ref{fig:bode}.(g)-(i) shows the mean and standard error for each experimental group. Similarly to \cite[Fig.~6]{CannonZahalak1982JB}, the phase shift in each experimental group changes very little across a wide range of frequencies before it reaches the second order zero at the natural frequency $\omega_{h}$ of $S_h (s)$.

Tab.~\ref{phase-shift} shows the mean and standard error for the phase shift of $S_{h} (s)$ in each experimental group across different frequencies. The data for the last three frequencies is excluded from the calculation due to the effect of the second order zero at $\omega_{h}$.

\subsection{Model Comparison Results}

\begin{figure}[!tbp]
    \centering
    \scriptsize 
        \def\svgwidth{.49\textwidth}
    	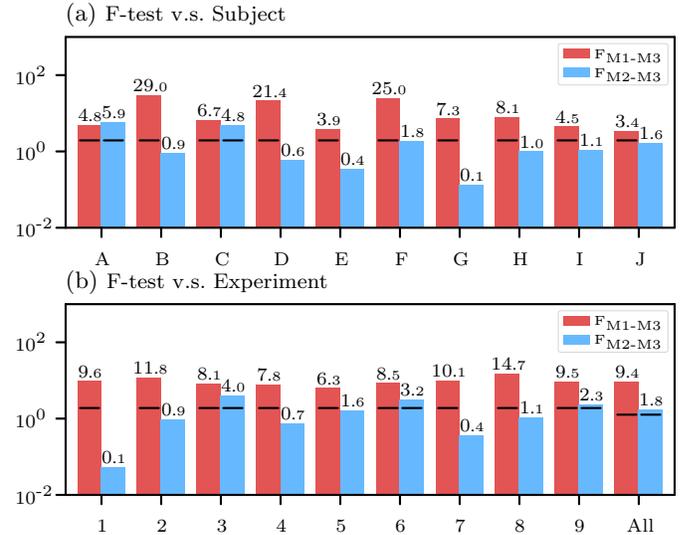
    \caption{Bar charts on log scale show first, all $\mathrm{F^{sub}_{Mi \mhyphen M3}}$ in (a), second all $\mathrm{F^{exp}_{Mi \mhyphen M3}}$ in the first nine columns of (b), and third $\mathrm{F^{all}_{Mi \mhyphen M3}}$ in the last column of (b), for $\mathrm{i = 1, 2}$. 
    The solid line appears on a bar if the F-statistic value is over the critical F-statistic value with a false-rejection probability of 0.05.}
    \label{fig:f-test}
\end{figure}

We now focus on the statistical significance analysis presented in Fig.~\ref{fig:f-test}.
Fig.~\ref{fig:f-test}.(a) shows a subject-wise comparison of the significance of the terms $B_{h} s$ and $H_{h} j$ that we use in M3. A critical F-statistic value of $1.95$ is calculated for $0.05$ false-rejection probability with $(9, 144)$ degrees of freedom. The results show that the values of $\mathrm{F^{sub}_{M1 \mhyphen M3}}$ for all subjects are higher than the critical F-statistic value. In particular, the values of $\mathrm{F^{sub}_{M1 \mhyphen M3}}$ for subjects B, D, and F exceed $20$. These results prove that the existence of $H_{h} j$ in M3 significantly improves modeling accuracy of $S_{h \mhyphen e \mslash \alpha}$ for all subjects. The values of $\mathrm{F^{sub}_{M2 \mhyphen M3}}$ are mostly below the critical F-statistic value except for subjects A and C. Another observation is that the value of $\mathrm{F^{sub}_{M2 \mhyphen M3}}$ is lower than the value of $\mathrm{F^{sub}_{M1 \mhyphen M3}}$ for most of the subjects except for subject A. 

Fig.~\ref{fig:f-test}.(b) shows an experiment-wise comparison of the models. A critical F-statistic value of $1.89$ is calculated for $0.05$ false-rejection probability with $(10, 160)$ degrees of freedom. The results show that the values of $\mathrm{F^{exp}_{M1 \mhyphen M3}}$ for all subjects are much higher than the critical F-statistic value. These results prove that the existence of $H_{h} j$ in M3 significantly improves modeling accuracy of $S_{h \mhyphen e \mslash \alpha}$ for all stiffness and inertia settings. The values of $\mathrm{F^{exp}_{M2 \mhyphen M3}}$ are mostly lower than the critical F-statistic value except for the three experiments with amplification factor $\alpha=4$ (Exp.~$3$, $6$, $9$). Also, we can see a clear increment of the values of $\mathrm{F^{exp}_{M2 \mhyphen M3}}$ (i.e. the significance of $B_{h} s$ in M3) as $\alpha$ gets higher. 

Regarding the significance of the terms $B_{h} s$ and $H_{h} j$ used in M3 over all subjects and all experiments, a critical F-statistic value of $1.27$ is calculated for a $0.05$ false-rejection probability with $(90, 1440)$ degrees of freedom. The value of $\mathrm{F^{all}_{M1 \mhyphen M3}}$ is much larger than $1.27$ while the value of $\mathrm{F^{all}_{M2 \mhyphen M3}}$ is only slightly above $1.27$. Although the effect of $B_{h} s$ cannot be completely ignored based on the results of these F-tests, we can claim that the term $H_{h} j$ has much more significance than the term $B_{h} s$ as used in M3.

\subsection{Complex Stiffness Results}

\setlength{\tabcolsep}{0.0pt}
\begin{table} [!tbp]
\caption{Observed Phase Shifts}
\centering
\scriptsize
\begin{tabular}{L{2.25cm} C{1.1cm} C{1.1cm} C{1.1cm} C{1.1cm} C{1.1cm} C{1.1cm}} \toprule
\multirow{2}{*}{$\mathrm{Subject}$} & \multicolumn{2}{c}{$\mathrm{Exp. \ 1 \mhyphen 3 \ (deg)}$} & \multicolumn{2}{c}{$\mathrm{Exp. \ 4 \mhyphen 6 \ (deg)}$} & \multicolumn{2}{c}{$\mathrm{Exp. \ 7 \mhyphen 9 \ (deg)}$} \\  [.5ex] 
& $\mathrm{Mean}$ & $\mathrm{S.E.}$ & $\mathrm{Mean}$ & $\mathrm{S.E.}$ & $\mathrm{Mean}$ & $\mathrm{S.E.}$ \\ [.0ex] 
\midrule
$\mathrm{A}$ & $27.8$ & $3.1$ & $25.4$ & $2.8$ & $18.1$ & $2.5$ \\
$\mathrm{B}$ & $27.2$ & $2.4$ & $34.8$ & $2.5$ & $35.2$ & $2.0$ \\
$\mathrm{C}$ & $16.7$ & $2.6$ & $21.6$ & $3.3$ & $22.8$ & $2.7$ \\
$\mathrm{D}$ & $34.7$ & $2.7$ & $38.5$ & $2.4$ & $33.3$ & $3.0$ \\
$\mathrm{E}$ & $17.6$ & $2.5$ & $10.7$ & $2.8$ & $11.1$ & $2.2$ \\
$\mathrm{F}$ & $33.7$ & $2.1$ & $33.4$ & $2.2$ & $27.5$ & $3.0$ \\
$\mathrm{G}$ & $23.9$ & $3.2$ & $16.7$ & $3.2$ & $15.2$ & $2.6$ \\
$\mathrm{H}$ & $19.3$ & $3.1$ & $23.4$ & $4.4$ & $25.3$ & $2.2$ \\
$\mathrm{I}$ & $18.3$ & $2.5$ & $14.5$ & $3.7$ & $10.4$ & $2.4$ \\
$\mathrm{J}$ & $19.9$ & $3.2$ & $20.0$ & $3.6$ & $14.6$ & $4.1$ \\ \midrule
$\mathrm{Cuff}$ & $ 6.0$ & $0.6$ & $ 6.5$ & $0.9$ & $ 7.6$ & $1.2$ \\ [0ex]
\bottomrule
\end{tabular} \label{phase-shift}
\end{table}

\setlength{\tabcolsep}{0.0pt}
\begin{table*} [!tbp]
\caption{Complex Stiffness Parameters for M2}
\centering
\tiny
\begin{tabular}{L{1.34cm} C{1.92cm} C{1.34cm} C{1.34cm} C{1.34cm} C{1.34cm} C{1.34cm} C{1.34cm} C{1.34cm} C{1.34cm} C{1.34cm} C{1.34cm} C{1.30cm}} \toprule
 \multirow{2}{*}{\scriptsize $\mathrm{Exp}$} & \multirow{2}{*}{\scriptsize $\mathrm{Parameter}$} & \multicolumn{11}{c}{\scriptsize $\mathrm{Subject}$} \\ [1.2ex]
                      &                       & \scriptsize $\mathrm{A}$ & \scriptsize $\mathrm{B}$ & \scriptsize $\mathrm{C}$ & \scriptsize $\mathrm{D}$ & \scriptsize $\mathrm{E}$ & \scriptsize $\mathrm{F}$ & \scriptsize $\mathrm{G}$ & \scriptsize $\mathrm{H}$ & \scriptsize $\mathrm{I}$ & \scriptsize $\mathrm{J}$ & \scriptsize $\mathrm{Average}$ \\
 \midrule
 \multirow{3}{*}{\scriptsize $1$}   & $K_h \ \mathrm{(Nm / rad)}$ & $12.68$ & $28.67$ & $17.76$ & $16.88$ & $11.55$ & $13.41$ & $17.95$ & $17.37$ & $18.85$ & $13.77$ & $16.35$    \\
                        & $H_h \ \mathrm{(Nm / rad)}$ & $ 5.26$ & $14.09$ & $ 5.22$ & $12.38$ & $ 3.92$ & $ 7.26$ & $ 4.48$ & $ 5.22$ & $ 5.30$ & $ 2.56$ &  $5.80$    \\
                        &                       $R^2$ & $ 0.99$ & $ 0.98$ & $ 0.97$ & $ 0.98$ & $ 0.96$ & $ 1.00$ & $ 0.97$ & $ 0.95$ & $ 0.95$ & $ 0.96$ & $\mhyphen$ \\ \cmidrule[.5pt]{2-13}
 \multirow{3}{*}{\scriptsize $2$}   & $K_h \ \mathrm{(Nm / rad)}$ & $16.05$ & $21.43$ & $12.62$ & $16.85$ & $14.72$ & $ 9.49$ & $19.39$ & $14.62$ & $20.16$ & $10.78$ & $15.13$    \\ 
                        & $H_h \ \mathrm{(Nm / rad)}$ & $ 8.08$ & $11.57$ & $ 4.79$ & $ 9.58$ & $ 4.59$ & $ 5.98$ & $ 8.98$ & $ 5.42$ & $ 6.55$ & $ 4.43$ &  $6.64$    \\
                        &                       $R^2$ & $ 0.95$ & $ 0.99$ & $ 0.96$ & $ 0.98$ & $ 0.97$ & $ 0.98$ & $ 0.97$ & $ 0.96$ & $ 0.97$ & $ 0.96$ & $\mhyphen$ \\ \cmidrule[.5pt]{2-13}
 \multirow{3}{*}{\scriptsize $3$}   & $K_h \ \mathrm{(Nm / rad)}$ & $10.16$ & $18.59$ & $10.88$ & $10.60$ & $12.87$ & $ 8.83$ & $12.33$ & $12.41$ & $22.70$ & $10.03$ & $12.40$    \\
                        & $H_h \ \mathrm{(Nm / rad)}$ & $ 6.60$ & $ 9.70$ & $ 4.63$ & $ 7.19$ & $ 5.72$ & $ 7.04$ & $ 7.05$ & $ 5.14$ & $ 9.14$ & $ 4.35$ &  $6.44$    \\
                        &                       $R^2$ & $ 0.97$ & $ 0.98$ & $ 0.88$ & $ 0.90$ & $ 0.90$ & $ 0.99$ & $ 0.96$ & $ 0.95$ & $ 0.96$ & $ 0.97$ & $\mhyphen$ \\ \midrule
 \multirow{3}{*}{\scriptsize $4$}   & $K_h \ \mathrm{(Nm / rad)}$ & $28.69$ & $45.01$ & $30.81$ & $39.08$ & $35.16$ & $31.57$ & $41.56$ & $34.24$ & $62.06$ & $27.66$ & $36.52$    \\
                        & $H_h \ \mathrm{(Nm / rad)}$ & $ 9.95$ & $29.42$ & $14.01$ & $30.98$ & $ 7.13$ & $18.18$ & $13.72$ & $ 7.97$ & $15.69$ & $ 8.57$ & $13.75$    \\
                        &                       $R^2$ & $ 0.96$ & $ 0.96$ & $ 0.94$ & $ 0.98$ & $ 0.97$ & $ 0.98$ & $ 0.91$ & $ 0.96$ & $ 0.96$ & $ 0.96$ & $\mhyphen$ \\ \cmidrule[.5pt]{2-13}
 \multirow{3}{*}{\scriptsize $5$}   & $K_h \ \mathrm{(Nm / rad)}$ & $26.97$ & $32.64$ & $18.81$ & $27.25$ & $36.50$ & $23.94$ & $41.65$ & $47.88$ & $36.65$ & $17.92$ & $29.60$    \\
                        & $H_h \ \mathrm{(Nm / rad)}$ & $15.25$ & $21.69$ & $ 7.51$ & $21.78$ & $ 9.74$ & $16.28$ & $13.54$ & $23.64$ & $10.60$ & $ 7.26$ & $13.57$    \\
                        &                       $R^2$ & $ 0.93$ & $ 0.96$ & $ 0.95$ & $ 0.96$ & $ 0.93$ & $ 0.99$ & $ 0.89$ & $ 0.94$ & $ 0.93$ & $ 0.97$ & $\mhyphen$ \\ \cmidrule[.5pt]{2-13}
 \multirow{3}{*}{\scriptsize $6$}   & $K_h \ \mathrm{(Nm / rad)}$ & $24.23$ & $32.94$ & $25.85$ & $29.05$ & $26.09$ & $20.93$ & $24.41$ & $24.21$ & $26.29$ & $14.55$ & $24.37$    \\
                        & $H_h \ \mathrm{(Nm / rad)}$ & $14.23$ & $20.52$ & $11.31$ & $18.23$ & $ 8.48$ & $14.48$ & $ 9.93$ & $10.15$ & $12.11$ & $ 7.01$ & $12.03$    \\
                        &                       $R^2$ & $ 0.90$ & $ 0.97$ & $ 0.97$ & $ 0.98$ & $ 0.95$ & $ 0.99$ & $ 0.92$ & $ 0.97$ & $ 0.88$ & $ 0.95$ & $\mhyphen$ \\ \midrule
 \multirow{3}{*}{\scriptsize $7$}   & $K_h \ \mathrm{(Nm / rad)}$ & $45.12$ & $73.23$ & $66.65$ & $54.75$ & $63.99$ & $63.31$ & $78.08$ & $55.26$ &$108.33$ & $60.11$ & $65.12$    \\
                        & $H_h \ \mathrm{(Nm / rad)}$ & $13.52$ & $49.62$ & $34.33$ & $45.87$ & $11.70$ & $32.12$ & $20.20$ & $25.01$ & $23.61$ & $12.83$ & $23.90$    \\
                        &                       $R^2$ & $ 0.94$ & $ 0.98$ & $ 0.96$ & $ 0.94$ & $ 0.92$ & $ 0.97$ & $ 0.97$ & $ 0.97$ & $ 0.98$ & $ 0.95$ & $\mhyphen$ \\ \cmidrule[.5pt]{2-13}
 \multirow{3}{*}{\scriptsize $8$}   & $K_h \ \mathrm{(Nm / rad)}$ & $52.45$ & $55.48$ & $63.19$ & $59.58$ & $59.37$ & $41.71$ & $67.47$ & $69.80$ & $68.61$ & $41.52$ & $57.03$    \\
                        & $H_h \ \mathrm{(Nm / rad)}$ & $17.52$ & $43.15$ & $31.05$ & $39.03$ & $16.03$ & $21.37$ & $19.20$ & $28.97$ & $14.61$ & $15.99$ & $22.97$    \\
                        &                       $R^2$ & $ 0.91$ & $ 0.98$ & $ 0.97$ & $ 0.98$ & $ 0.95$ & $ 0.99$ & $ 0.97$ & $ 0.97$ & $ 0.94$ & $ 0.97$ & $\mhyphen$ \\ \cmidrule[.5pt]{2-13}
 \multirow{3}{*}{\scriptsize $9$}   & $K_h \ \mathrm{(Nm / rad)}$ & $40.87$ & $65.45$ & $45.33$ & $46.37$ & $46.67$ & $39.03$ & $49.83$ & $63.44$ & $67.79$ & $33.68$ & $48.63$    \\
                        & $H_h \ \mathrm{(Nm / rad)}$ & $20.08$ & $38.08$ & $17.76$ & $24.44$ & $14.52$ & $22.11$ & $17.62$ & $33.32$ & $18.52$ & $13.92$ & $20.93$    \\
                        &                       $R^2$ & $ 0.93$ & $ 0.98$ & $ 0.96$ & $ 0.98$ & $ 0.92$ & $ 0.95$ & $ 0.92$ & $ 0.97$ & $ 0.95$ & $ 0.93$ & $\mhyphen$ \\ \midrule
                        & $\beta_0$ &  $0.03$ & $-0.55$ & $-0.55$ & $-0.21$ & $-0.11$ & $-0.01$ & $ 0.00$ & $-0.56$ & $-0.10$ & $-0.26$ & $-0.23$ \\ [0.3ex]
 \scriptsize $\mathrm{Power \ Law}$  & $\beta_1$ &  $0.73$ &  $1.21$ &  $1.12$ &  $1.03$ &  $0.70$ &  $0.85$ &  $0.70$ &  $1.10$ &  $0.73$ &  $0.84$ &  $0.90$ \\ [0.3ex]
                        &     $R^2$ &  $0.77$ &  $0.97$ &  $0.97$ &  $0.94$ &  $0.86$ &  $0.96$ &  $0.83$ &  $0.91$ &  $0.87$ &  $0.73$ &  $0.95$ \\
\bottomrule
\end{tabular}
\label{m2-parameter}
\end{table*}

\begin{figure*}[!tbp]
    \centering
    \tiny
        \def\svgwidth{1.\textwidth}
    	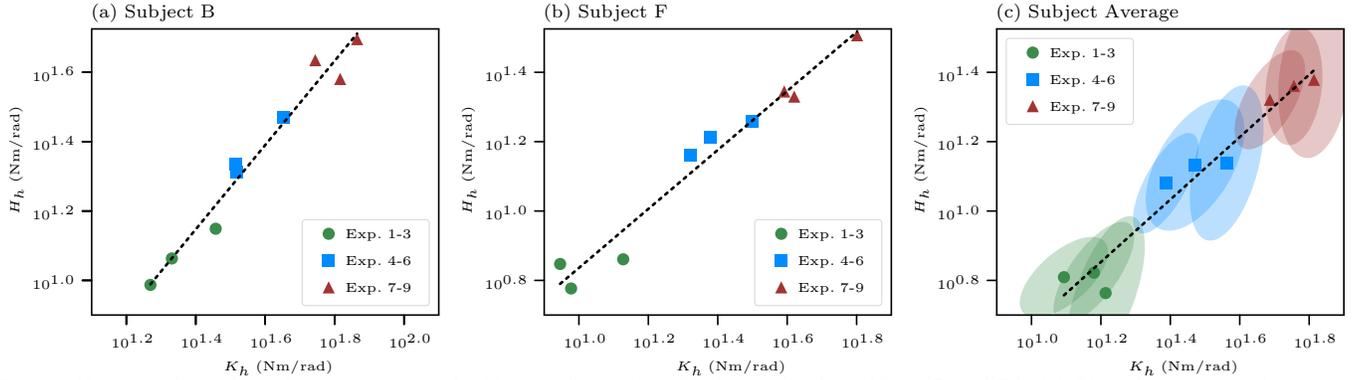
    \caption{Here we show plots of $H_{h}$ versus $K_{h}$ for M2 on a log scale, with the results for subjects B and F in (a)-(b) and the geometric average across all subjects in (c). The dash lines and the ellipsoids show the linear regression results and the co-variances on the log scale.}
    \label{fig:regression}
\end{figure*}

In our single-subject pilot study \cite{HeHuangThomasSentis2019IROS}, a linear regression is applied to describe the relationship between $H_{h}$ and $K_{h}$ identified from M2 and M3. However, based on our frequency domain results, the phase shifts of $S_{h \mhyphen e \mslash \alpha} (s)$ for some subjects are not consistent over different stiffness values. Therefore, a linear relationship between $H_{h}$ and $K_{h}$ is not always ensured for all subjects. Instead, we apply linear regression between the base $10$ logarithms of $H_{h}$ and $K_{h}$ and use it to identify a power law between these two parameters. Since the value of $H_{h}$ is not guaranteed to be positive from the parameter identification of M3, we only calculate the power law between $H_{h}$ and $K_{h}$ of M2.

Tab.~\ref{m2-parameter} shows the identified parameter values of $H_{h}$ and $K_{h}$ using M2, with a coefficient of determination ($R^2$) in the range of $0.88 \sim 1.00$.
We define $\beta_0$ and $\beta_1$ as the intercept and slope of the linear regression equation between the base $10$ logarithms of $H_{h}$ and $K_{h}$. 
From the parameter identification results using M2, a very strong linear relationship between logarithms is observable across all subjects, with an $R^2$ value in the range of $0.73 \sim 0.96$. 
Fig.~\ref{fig:regression}.(a)-(b) show the regression results of subjects B and F.

The last three rows of Tab.~\ref{m2-parameter} show the identified power law parameters.
The damping ratio and low-frequency phase shift of M2 can be expressed as 
\begin{flalign}
\zeta_{h \mhyphen e \mslash \alpha} & = c_{h} / 2, \ \
\phi_{h \mhyphen e \mslash \alpha} = \tan^{-1}(c_{h}), \label{eq:zt-m2}
\end{flalign}
where $c_h$ is a hysteretic damping loss factor \cite{BishopJohnson2011book} expressed as
\begin{flalign}
c_{h} & \triangleq H_{h} / K_{h} = 10 ^{\beta_0} \cdot K_{h} ^ {\beta_1 - 1}  \label{eq:ch}
\end{flalign}
obtained by substituting the power law $H_{h} = 10 ^{\beta_0} \cdot K_{h} ^ {\beta_1}$.


The last column of Tab.~\ref{m2-parameter} shows the geometric average (i.e. arithmetic average of the logarithms) of the complex stiffness parameters across all subjects. 
We apply a linear regression to the logarithms of these average values and identify a power law of $\beta_0 = -0.23$, $\beta_1 = 0.90$, and $R^2 = 0.95$ (Fig.~\ref{fig:regression}.(c)). 
As the subject average stiffness increases from $12.40$ to $65.12$ $\mathrm{Nm/rad}$, the value of $\zeta_{h \mhyphen e \mslash \alpha}$ decreases from $0.23$ to $0.19$ as calculated using \eqref{eq:zt-m2} and is within a 1-standard deviation range of the damping ratio of the minimally damped mode of the human arm ($0.26 \pm 0.08$) described in \cite{PerreaultKirschCrago2004EBR}.

As in \cite{HeHuangThomasSentis2019IROS}, the correlation between $H_{h}$ and $K_{h}$ can be introduced into M2 to reduce it to a 1-parameter complex stiffness model. 
Adopting this reduced model allows simplifying \eqref{Sh2} to
\begin{equation} \label{eq:Sh2-new}
S_{h} (s) = \tau_{c}/\theta_{e} = M_{h} s^2 + K_{h}(1 + c_{h} j),
\end{equation}
and the dynamic stiffness of the human coupled with the exoskeleton $S_{h \mhyphen e} (s)$ becomes,
\begin{equation} \label{eq:Sh-e2-new}
S_{h \mhyphen e} (s) = \tau_{s}/\theta_{e} = M_{h \mhyphen e} s^2 + K_{h}(1 + c_{h} j),
\end{equation}
where $M_{h \mhyphen e} = M_{h} + M_{e}$ is the combined inertia between the human and the exoskeleton. 
Similarly to \eqref{eq:zt-m2}, the damping ratio and low-frequency phase shift of $S_{h} (s)$ and $S_{h \mhyphen e} (s)$ can also be expressed as $c_h / 2$ and $\tan ^ {-1} (c_h)$.


\section{Loop Shaping Methods} 
The amplification feedback we discuss in this section is the same as the direct amplification feedback shown in Fig.~\ref{block-diagram} in which the amplification command is $-\tau_c$ multiplied by $\alpha -1$. 
But instead of a constant value of $\alpha$ across all frequencies, we introduce a frequency dependent amplification transfer function $\alpha (s) = \mathrm{k_p} \cdot F (s) + 1$,
where $\mathrm{k_p}$ is a proportional gain and $F (s)$ is a fractional order controller customized according to the complex stiffness behavior displayed by  users. 

\subsection{Proportional Amplification}

Based on \eqref{eq:Sh2-new} and \eqref{eq:Sh-e2-new}, the plant transfer function $P (s)$ from $\tau_d$ to $\tau_c$ can be expressed as
\begin{equation}
P (s) = \frac{\text{\small $S_{h} (s)$}}{\text{\small $S_{h \mhyphen e} (s)$}} \cdot \text{\small $G_{SEA} (s)$} = \frac{\text{\small $M_{h} s^2 + K_{h}(1 + c_{h} j)$}}{\text{\small $M_{h \mhyphen e} s^2 + K_{h}(1 + c_{h} j)$}} \cdot \text{\small $G_{SEA} (s)$}
\end{equation}
where the SEA transfer function $G_{SEA} (s) = \tau_s / \tau_d$ acts as a 2nd order low-pass filter. Because of the high bandwidth of the SEA force controller, the natural frequency $\omega_{SEA}$ of $G_{SEA} (s)$ is much greater than the natural frequencies $\omega_{h \mhyphen e} = \sqrt{K_{h} / M_{h \mhyphen e}}$ and $\omega_{h} = \sqrt{K_{h} / M_{h}}$ of $S_{h \mhyphen e} (s)$ and $S_{h} (s)$.

Considering the frequency domain properties from low to high frequencies, $P (s)$ has a pair of conjugate poles at $\omega_{h \mhyphen e}$, then a pair of conjugate zeros at $\omega_{h}$ and then another pair of conjugate poles at $\omega_{SEA}$ (Fig.~\ref{fig:concept}). 
Between $\omega_{h \mhyphen e}$ and $\omega_{h}$, $S_{h \mhyphen e} (s)$ is dominated by its inertia effect and the magnitude of $P (s)$ decreases while the phase decreases from $0 \degree$. On the other hand, $S_{h} (s)$ is still dominated by the complex stiffness and prevents the phase moving below $\tan^{-1}(c_{h}) - 180\degree$. 

If we apply a very large value of $\mathrm{k_p}$, the gain crossover of $P (s)$ falls beyond $\omega_{SEA}$. 
The phase margin with such crossover is very close to zero because of the $\mathrm{2nd}$ order SEA dynamics. 
Also, the closed loop behavior amplifies the high frequency sensor noise from the actual signal from $\tau_c$ (which is usually de-noised by a low-pass filter beyond the frequency of $\omega_{SEA}$ that makes the closed loop even more likely to be unstable). 
Similarly to \cite{HeThomasPaineSentis2019ACC}, the crossover frequency cannot be placed between $\omega_{h}$ and $\omega_{SEA}$ because multiple crossovers could easily occur. 
Besides the multiple crossovers, this frequency range is also outside of the tested frequency ranges of Exp.~$1$-$9$. 
The unmodeled dynamics from the human and cuff will cause additional stability issues if a crossover is placed there. 
Instead, a new crossover can be safely placed at the frequency between $\omega_{h \mhyphen e}$ and $\omega_{h}$ using a smaller $\mathrm{k_p}$ (Fig.~\ref{fig:concept}). 
As a rule of thumb, $\mathrm{k_p}$ can be set as 
\begin{equation} \label{kp}
\begin{aligned}
\mathrm{k_p} & = (\omega_{\mathrm {gc}} / \omega_{h}) ^ {2} = (M_{h \mhyphen e} / M_{h}) ^ {\frac{1}{2}},  \\
\omega_{\mathrm {gc}} & \triangleq \sqrt{K_h / (M_{h \mhyphen e} \cdot M_{h}) ^ {\frac{1}{2}}},
\end{aligned}
\end{equation}
where the crossover $\omega_{\mathrm {gc}}$ of $\mathrm{k_p} \cdot P(s)$ is exactly in the middle between $\omega_{h \mhyphen e}$ to $\omega_{h}$ in the log scale.

\subsection{Fractional Order Amplification}


In \cite{HeThomasPaineSentis2019ACC}, an additional integral term is added to the proportional gain $\mathrm{k_p}$ to boost the amplification at low frequencies while maintaining the same crossover frequency between $\omega_{h \mhyphen e}$ and $\omega_{h}$. However, a PI controller has a $-90 \degree$ phase at low frequency, which can result in loss of stability if the zero of the PI controller is too close to the crossover.

In \cite{HeHuangThomasSentis2019IROS}, a fractional order controller was proposed to take advantage of the complex stiffness model,
\begin{equation} \label{F}
F (s) = \mathrm{k_f} \cdot s ^ {- \mathrm f},
\end{equation}
where $\mathrm f$ is the fractional order (that is, a non-integer power of $s$) of $F (s)$ and $\mathrm{k_f}$ is a gain which allows tuning the magnitude of $F (s)$ in the frequency domain. The fractional order controller in \eqref{F} has its magnitude decreasing $-20\cdot \mathrm f$ dB per decade and its phase staying at $-90\cdot \mathrm f$ degrees at all frequencies. 

\begin{figure}[!tbp]
\centering
\small
\def\svgwidth{.49\textwidth}
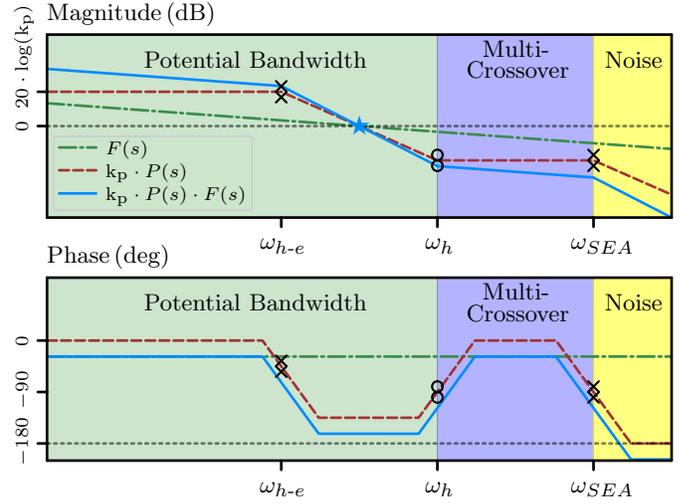
\caption{These conceptual bode plots show $P (s)$ with its poles (crosses) and zeros (circles). The various regions are color-coded: the model is trustworthy in the green region, the blue region reflects the multi-crossover behavior which makes an amplification controller design unreliable, and the yellow region is dominated by sensor noise from $\tau_c$. 
A fractional-order controller $F (s)$ complements a proportional controller $\mathrm{k_p}$ by boosting the low-frequency amplification.}
\label{fig:concept}
\end{figure}

\begin{figure*}[!tbp]
    \centering
    \tiny
        \def\svgwidth{1.\textwidth}
    	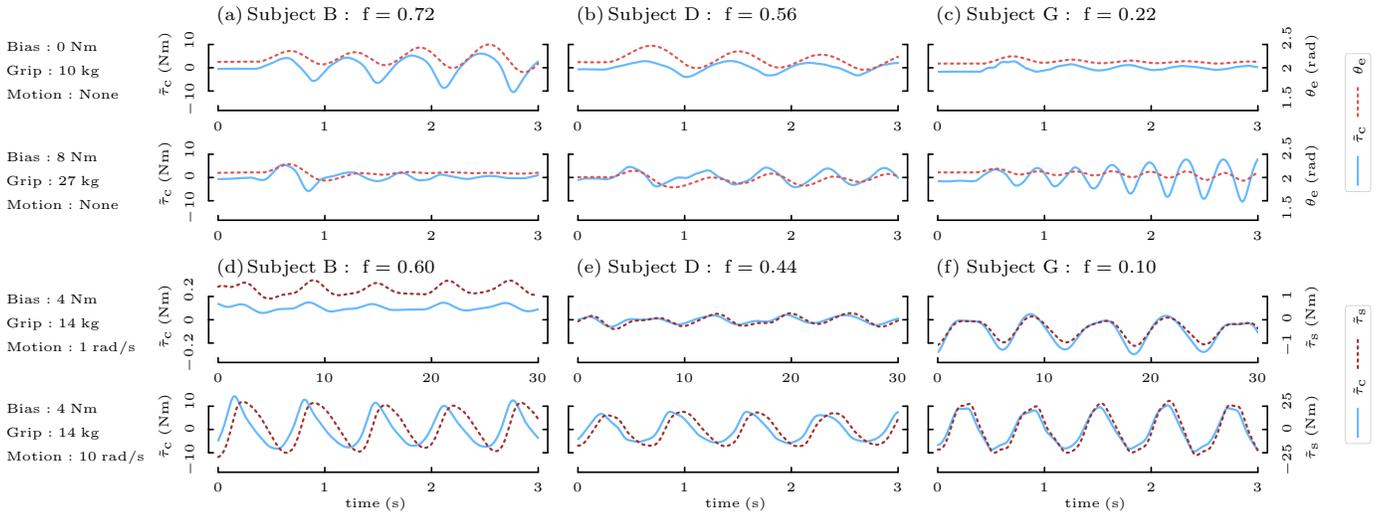
    \caption{(a)-(c) show the responses of $\theta_e$ (dash red) and $\tilde{\tau}_c$ (solid blue) for the post-tuning tests on subjects B, D and G. (d)-(f) show the responses of $\tilde{\tau}_s$ (dash red) and $\tilde{\tau}_c$ (solid blue) for the amplification tests on subjects B, D and G.}
    \label{fig:loop-shaping-results}
\end{figure*}

By multiplying \eqref{F} by the proportional gain $\mathrm{k_p}$, the gain of the controller is increased at low frequency and reduced at high frequency (for further de-noising the measurement of $\tau_c$).
If $\mathrm{k_f}$ is tuned to make $F(s)$ have the exact same crossover frequency as $\mathrm{k_p} \cdot P(s)$, we will obtain the magnitude bode plot $\mathrm{k_p} \cdot P(s) \cdot F(s)$ rotated from $\mathrm{k_p} \cdot P(s)$ with pivot at the point of the gain crossover frequency (Fig.~\ref{fig:concept}).
Since the exact crossover frequency of $\mathrm{k_p} \cdot P(s)$ varies with the value of $K_h$, $\mathrm{k_f}$ can be set as
\begin{equation} \label{kf}
\mathrm{k_f} = \hat{\omega}_{\mathrm {gc}} ^ {\mathrm f},  \quad
\hat{\omega}_{\mathrm {gc}} \triangleq \sqrt{\hat{K}_h / (M_{h \mhyphen e} \cdot M_{h}) ^ {\frac{1}{2}}},
\end{equation}
where $\hat{\omega}_{\mathrm {gc}}$ is chosen as a nominal crossover frequency of $\mathrm{k_p} \cdot P(s)$ with $\hat{K}_h \triangleq (\ubar{K}_h \cdot \bar{K}_h) ^ {1/2}$ being the geometric mean between the lower bound $\ubar{K}_h$ and the upper bound $\bar{K}_h$.

Because of the non zero phase shift associated with the complex stiffness behavior, a positive phase margin can be guaranteed if $0 < \mathrm f < \tan^{-1}(c_{h}) / 90$.
The fractional order controller can be precisely designed for all subjects based on the values of $\beta_0$ and $\beta_1$ shown in Tab.~\ref{m2-parameter} through the settings
\begin{equation} \label{f}
\mathrm f = 
\begin{cases}
\tan ^ {-1} (10^{\beta_0} \cdot  \bar{K}_h ^ {\beta_1 - 1}) / 90 - \phi / 90, &  \mathrm{if}\ \beta_1 < 1, \\
\tan ^ {-1} (10^{\beta_0} \cdot \ubar{K}_h ^ {\beta_1 - 1}) / 90 - \phi / 90, &  \mathrm{if}\ \beta_1 \geq 1,
\end{cases}
\end{equation}
where $\phi > 0$ is a user-defined guaranteed phase margin. 
Differently from \cite{HeHuangThomasSentis2019IROS} where a constant $c_h$ is assumed, \eqref{f} considers the lowest value of $c_h$ of a subject in the stiffness range $[\ubar{K}_h, \, \bar{K}_h]$.

As a fractional-order controller, $F(s)$ cannot be implemented directly into a computational control process. However, from \cite{HeHuangThomasSentis2019IROS}, we can approximate it as the product of many 1st order lag filters,
\begin{flalign} \label{Fs}
F (s) & = \frac{\mathrm{k_f}  \hfill}{p_1 ^ {\mathrm f}} \cdot \prod_{i=1}^{n} \frac{1 + s/ \, z_i \hfill}{1 + s/p_i}, \\
z_i / p_i & = r_{zp}, \quad \mathrm{for} \; \; i = 1, \, 2, \, \cdots, \, n \\
p_i / p_{i-1} & = r_{pp}, \quad \mathrm{for} \; \; i = 2, \, 3, \, \cdots, \, n,
\end{flalign}
where $n$ is the number of lag filters and the pole and the zero for each lag filter are $- p_i$ and $- z_i$. We define $r_{zp}$ such that all lag filters have an equal distance between the pole and the zero, and we define $r_{pp}$ such that there is a constant distance between adjacent lag filters (in $\log$ frequency space). The amplification controller in \eqref{Fs} functions as a fractional-order filter in the frequency range of $[p_1, \; z_n]$ $\mathrm{rad/s}$. The fractional order can be approximated as $\mathrm f \approx \log(r_{zp}) / \log(r_{pp})$.

\subsection{Experimental Protocol for Loop Shaping} \label{trigger}

Based on \eqref{f}, 
we conducted loop shaping experiments on subjects B, D and G who, respectively, had the highest value, the closest value to $1$, and the lowest value of $\beta_1$ across all subjects. Our loop shaping study consists of two tuning experiments and two amplification experiments. 

The value of $M_e$ is $1.01$ $\mathrm{kg \cdot m^2}$, which includes a $4.5$ $\mathrm{kg}$ load at the end of the exoskeleton arm. Although we do not measure the value of $M_h$ directly from our subjects, an average $M_h$ of $0.11$  $\mathrm{kg \cdot m^2}$ can be obtained from a $10$-subject measurement study presented in \cite{CannonZahalak1982JB}. Based on these inertia values and \eqref{kp}, we set $\mathrm{k_p} = 3.2$.

As shown in Tab.~\ref{m2-parameter}, the human stiffness changes from $10.03$ to $108.33$ $\mathrm{Nm/rad}$ across all subjects and all experiments, which gives us a nominal value of $\hat{K}_h = 32.96 \ \mathrm{Nm/rad}$. Based on \eqref{kf} and \eqref{Fs}, we compute $\hat{\omega}_{\mathrm{gc}} \approx 10 \ \mathrm{rad/s}$ and implement an approximate fractional-order controller using $5$ lag filters with $p_1 = 1$ and $r_{pp} = 10 ^ {0.5}$  
such that $\hat{\omega}_{\mathrm{gc}}$ is located at the center of the frequency range defined by $[p_1, \, p_5]$ $\mathrm{rad/s}$. 

The two tuning experiments we perform aim to find out the fractional order of a subject where the minimum phase margin $\phi$ is near zero. From \eqref{f}, we gradually increase the fractional order, $\mathrm{f}$, from low value to higher values until the exoskeleton starts to oscillate. We do that with subjects employing low and high human stiffness behaviors. The maximum stable value of $\mathrm f$ will be the lower value between the two stiffness cases. An important advantage is that this tuning strategy does not require prior knowledge of the human complex stiffness. Similarly to the modeling experiments previously presented, we regulate the low and high stiffness of a subject by setting the gripping force as $10$ and $27$ $\mathrm{kg}$ and the bias torque as $0$ and $8$ $\mathrm{Nm}$.


After the tuning experiments outlined above, we subtract $0.12$ from the marginally stable fractional order, which provides a minimum phase margin $\phi = 10.8 \degree$. Then, we conduct two amplification experiments both with a gripping force of $14$ $\mathrm{kg}$ and a bias torque of $4$ $\mathrm{Nm}$. These two experiments are conducted using sinusoidal voluntary movements performed by the subjects with frequencies of $1$ and $10$ $\mathrm{rad/s}$. The voluntary sinusoidal movements are guided by showing the subject a visual signal of the actual joint position $\theta_e$ and the desired sinusoidal wave on a screen. The amplification factor $\alpha$ for these sinusoidal voluntary movements can be calculated from the experimental data after the experiments.

\section{Loop Shaping Results}

In order to study the performance of the tuned amplification controllers for various subjects, we define $\tilde{\tau}_s \triangleq \tau_s + \tau_g - \mathrm{bias}$ and $\tilde{\tau}_c \triangleq \tau_c - \mathrm{bias}$. 
The real-time amplification factor for the proposed amplification controller can be expressed as $\alpha (t) = \tilde{\tau}_s (t) / \tilde{\tau}_c (t) + 1$.

\subsection{Tuning Results} \label{ssec:tun_res}
After gradually increasing the fractional order for low and high human stiffness behaviors until the exoskeleton starts to oscillate continuously, we obtain the values $\mathrm{f} = 0.72, \, 0.56, \, 0.22$ for subjects B, D, G.
In order to display the tuning results concisely, we conduct two post-tuning tests involving low and high human stiffness setups.
We attach a set of mechanical springs to the tip of the exoskeleton arm (Fig.~\ref{setup}.(b)) and quickly detach it to test the dynamic response of the controller. The response of $\theta_e$ and $\tilde{\tau}_c$ for the post-tuning tests are shown in Fig.~\ref{fig:loop-shaping-results}.(a)-(c).

In Tab.~\ref{m2-parameter}, we had identified that $\beta_1$ for subject B was greater than $1$. This explains why the post-tuning test for subject B applying high stiffness is less oscillatory than the post-tuning test applying low stiffness.
Because the value of $\beta_1$ is very close to $1$ for subject D, the results for both post-tuning tests are very similar. 
Similarly, the high stiffness post-tuning test for subject G is more oscillatory than the low stiffness post-tuning test because $\beta_1 < 1$.

\subsection{Amplification Results}\label{ssec:aug_res}

When we subtract $0.12$ from $\mathrm f$ for subjects B, D and G, we get the values $0.60$, $0.44$ and $0.10$.
The behaviors of $\tilde{\tau}_s$ and $\tilde{\tau}_c$ for sinusoidal voluntary movements between $1$ and $10$ $\mathrm{rad/s}$ demonstrate that the exoskeleton is stable for all subjects using our proposed custom robust amplification controllers (Fig.~\ref{fig:loop-shaping-results}.(d)-(f)). 
The values for the gain and the phase shift for $\tilde{\tau}_s$ and $\tilde{\tau}_c$ during the amplification tests are shown in Tab.~\ref{tab:amplification-factor}.

Notice that the subjects are able to maintain values between $| \frac{\tilde{\tau}_s}{\tilde{\tau}_c}| = 2.83 \sim 2.99$ with a voluntary motion of $10$ $\mathrm{rad/s}$. In our prior research which did not incorporate the proposed complex stiffness model \cite{HeThomasPaineSentis2019ACC}, the value of $| \frac{\tilde{\tau}_s}{\tilde{\tau}_c}|$ was between $1.46 \sim 1.58$ at $6.3$ $\mathrm{rad/s}$ (experimentally validated), and a value of $1.12$ at $10$ $\mathrm{rad/s}$ (theoretically estimated). Therefore, our proposed control strategy shows a $81 \sim 88 \%$ improvement in the magnitude when using a dynamical amplification factor $\alpha (s) = \tilde{\tau}_s (s) / \tilde{\tau}_c (s) + 1$.

\section{Discussion}

The sensor configuration for this experiment measures the deflection at the exoskeleton's hinge joint as well as the human torque using the cuff's six-axis force/torque sensor. Using this setup, we conduct a test of superposition for differentiating between the human elbow and cuff impedances. Fig.~\ref{fig:bode}.(g)-(i) and Tab.~\ref{phase-shift} show the frequency data and phase shift values of the cuff attached to a rigid object. The phase of the cuff is lower than the phases of all human subjects. Furthermore, the magnitude of the cuff stiffness is above $60$ $\mathrm{dB}$ ($1000$ $\mathrm{Nm / rad}$), which is significantly higher than the stiffness values of all human subjects as shown in Tab.~\ref{phase-shift}. We conclude that the human impedance becomes the dominant factor measured in our experiments.

While the noticeable phase shift values observed in this study are consistent with the ankle and elbow joint phase values reported in \cite{AgarwalGottlieb1977JBE, GottliebAgarwal1978JB, ZahalakHeyman1979JBE, CannonZahalak1982JB}, some research studies also report very small phase shift values for other human joints. 
Yet, our proposed complex stiffness model still holds for those results. 
For example, the damping ratio for the human knee joint reported in \cite[Fig.~5]{PopeCrowninshieldMillerJohnson1976JB} is $0.02$, which results in a phase shift of $2.34 \degree$ using equation \eqref{eq:zt-m2}. This kind of small hysteretic damping characteristic can be easily overlooked.

Low-frequency phase shifts are found in muscle spindles \cite{PoppeleBowman1970JN} and arteries \cite{WesterhofNoordergraaf1970JB} of mammals, suggesting that joint hysteretic damping could be due to the bio-mechanical properties of the human tissue. Therefore, we suspect that the human neuromuscular system, either through muscle and tendon hysteresis or through neural hysteretic behavior, is the mechanism behind our hysteretic damping hypothesis.



\setlength{\tabcolsep}{0.0pt}
\begin{table} [!tbp]
\caption{Observed Dynamical Amplification}
\scriptsize
\centering
\begin{tabular}{L{0.8cm} C{0.8cm} C{1.75cm} C{1.75cm} C{1.85cm} C{1.85cm}} \toprule
  \scriptsize Subject & \scriptsize $\mathrm{f}$ & \small $| \frac{\tilde{\tau}_s}{\tilde{\tau}_c}|$\tiny$\mathrm{(1 \, rad/s)}$ & \small $\angle \frac{\tilde{\tau}_s}{\tilde{\tau}_c}$\tiny$\mathrm{(1 \, rad/s)}$ & \small $| \frac{\tilde{\tau}_s}{\tilde{\tau}_c}|$\tiny$\mathrm{(10 \, rad/s)}$ & \small $\angle \frac{\tilde{\tau}_s}{\tilde{\tau}_c}$\tiny$\mathrm{(10 \, rad/s)}$  \\ [.5ex] 
 \midrule
 $\mathrm B$   & $0.60$  & $10.86$  & $-25.2 \degree$  & $2.83$  & $-54.9 \degree$ \\ 
 $\mathrm D$   & $0.44$  & $ 6.74$  & $-28.2 \degree$  & $2.84$  & $-42.5 \degree$ \\ 
 $\mathrm G$   & $0.10$  & $ 3.70$  & $ -8.8 \degree$  & $2.99$  & $-12.6 \degree$ \\ [0ex]
\bottomrule
\end{tabular} \label{tab:amplification-factor}
\end{table}

Because Fig.~\ref{fig:bode}.(g)-(i) shows a consistent phase shift across a wide range of low frequencies, it is natural to consider that the phase behavior of $S_h (s)$ has already reached a low-frequency asymptote at the lowest tested frequency and it will not change much at lower frequencies than that. This is difficult to experimentally verify because lower frequencies require longer experimental times making it harder for the subjects. Nonetheless, our lowest tested frequency, $2$ $\mathrm{rad/s}$ ($\approx 0.3$ $\mathrm{Hz}$), is below the frequencies reported in references \cite{AgarwalGottlieb1977JBE, GottliebAgarwal1978JB, ZahalakHeyman1979JBE, CannonZahalak1982JB, HunterKearney1982JB}.
In addition, our tested frequency range covers the frequencies that are important for practical control system design.

In this research, we use sinusoidal perturbations to identify a frequency domain model of human complex stiffness. 
Ref. \cite{AgarwalGottlieb1977JBE, GottliebAgarwal1978JB} use white noise perturbations for system identification, also revealing a non-zero phase shift at low frequencies and consistent damping ratios. Because our complex stiffness model is non-causal and nonlinear, it does not have an exact model representation in the time domain \cite{InaudiKelly1995JEM}. For this reason, it is difficult to identify the human complex stiffness behavior using impulse, step and ramp perturbations. 



Although the fractional order part of the proposed amplification controller is only useful for shaping the behavior in a certain frequency range, it is sufficient to demonstrate crossover at frequencies that could not be robustly stable with the conventional human joint model. Between this paper and \cite{HeThomasPaineSentis2019ACC}, we have a natural comparison between the control design problem with and without the hysteretic adjustment to the human impedance. 
The result is clear: without complex stiffness, controllers must be designed to cross over before the lowest natural frequency resulting from the human and exoskeleton inertia and the softest human stiffness; with the modification, the crossover can exceed this frequency by implementing an approximate fractional-order controller.


\section{Conclusion}
Exoskeletons with feedback human forces must be coupled stable with the natural human impedance to avoid undesired vibrations.
This paper presents a model for human impedance using an imaginary stiffness term to fill an energy-dissipation role similar to damping.
The paper also presents experiments which demonstrate that this new term is a more significant contributor to model accuracy than a linear damping term for cyclic motion of the elbow in the $10$-subject cohort we studied.
The loop shaping experiments demonstrate the stability and bandwidth of our controller and highlight the importance of testing both maximum and minimum human stiffness cases when tuning the fractional order controller. 





\bibliographystyle{IEEEtran}
\bibliography{main}

\begin{thebibliography}{10}
\providecommand{\url}[1]{#1}
\csname url@samestyle\endcsname
\providecommand{\newblock}{\relax}
\providecommand{\bibinfo}[2]{#2}
\providecommand{\BIBentrySTDinterwordspacing}{\spaceskip=0pt\relax}
\providecommand{\BIBentryALTinterwordstretchfactor}{4}
\providecommand{\BIBentryALTinterwordspacing}{\spaceskip=\fontdimen2\font plus
\BIBentryALTinterwordstretchfactor\fontdimen3\font minus
  \fontdimen4\font\relax}
\providecommand{\BIBforeignlanguage}[2]{{%
\expandafter\ifx\csname l@#1\endcsname\relax
\typeout{** WARNING: IEEEtran.bst: No hyphenation pattern has been}%
\typeout{** loaded for the language `#1'. Using the pattern for}%
\typeout{** the default language instead.}%
\else
\language=\csname l@#1\endcsname
\fi
#2}}
\providecommand{\BIBdecl}{\relax}
\BIBdecl

\bibitem{YagnNicolas1890PatentUSpatent}
N.~Yagn, ``Apparatus for facilitating walking, running, and jumping,'' \emph{US
  patent}, vol. 420179, 1890.

\bibitem{MakinsonBodineFitck1969report}
J.~B. Makinson, D.~P. Bodine, and B.~R. Fick, ``Machine augmentation of human
  strength and endurance {H}ardiman {I} prototype project,'' Specialty
  Materials Handling Products Operation, General Electric Company, Tech. Rep.,
  1969.

\bibitem{KazerooniGuo1993JDSMC}
H.~Kazerooni and J.~Guo, ``Human extenders,'' \emph{Journal of Dynamic Systems,
  Measurement, and Control}, vol. 115, no.~2B, pp. 281--290, 1993.

\bibitem{Kazerooni2005IROS}
H.~Kazerooni, ``Exoskeletons for human power augmentation,'' in
  \emph{Intelligent Robots and Systems (IROS), 2005 IEEE/RSJ International
  Conference on}.\hskip 1em plus 0.5em minus 0.4em\relax IEEE, 2005, pp.
  3459--3464.

\bibitem{DollarHerr2008TRO}
A.~M. Dollar and H.~Herr, ``Lower extremity exoskeletons and active orthoses:
  challenges and state-of-the-art,'' \emph{IEEE Transactions on Robotics},
  vol.~24, no.~1, pp. 144--158, 2008.

\bibitem{JacobsenOlivier2014Patent}
S.~C. Jacobsen and M.~X. Olivier, ``Contact displacement actuator system,''
  Sep.~30 2014, {U}S Patent 8,849,457.

\bibitem{FontanaVertechyMarcheschiSalsedoBergamasco2014RAM}
M.~Fontana, R.~Vertechy, S.~Marcheschi, F.~Salsedo, and M.~Bergamasco, ``The
  body extender: A full-body exoskeleton for the transport and handling of
  heavy loads,'' \emph{IEEE Robotics \& Automation Magazine}, vol.~21, no.~4,
  pp. 34--44, 2014.

\bibitem{ZhangFiersWitteJacksonPoggenseeAtkesonCollins2017Science}
J.~Zhang, P.~Fiers, K.~A. Witte, R.~W. Jackson, K.~L. Poggensee, C.~G. Atkeson,
  and S.~H. Collins, ``Human-in-the-loop optimization of exoskeleton assistance
  during walking,'' \emph{Science}, vol. 356, no. 6344, pp. 1280--1284, 2017.

\bibitem{LeeKimBakerLongKaravasMenardGalianaWalsh2018JNR}
S.~Lee, J.~Kim, L.~Baker, A.~Long, N.~Karavas, N.~Menard, I.~Galiana, and C.~J.
  Walsh, ``Autonomous multi-joint soft exosuit with augmentation-power-based
  control parameter tuning reduces energy cost of loaded walking,''
  \emph{Journal of Neuroengineering and Rehabilitation}, vol.~15, no.~1, p.~66,
  2018.

\bibitem{KongMoonJeonTomizuka2010TMech}
K.~Kong, H.~Moon, D.~Jeon, and M.~Tomizuka, ``Control of an exoskeleton for
  realization of aquatic therapy effects,'' \emph{IEEE/ASME Transactions on
  Mechatronics}, vol.~15, no.~2, pp. 191--200, 2010.

\bibitem{KimDeshpande2017IJRR}
B.~Kim and A.~D. Deshpande, ``An upper-body rehabilitation exoskeleton harmony
  with an anatomical shoulder mechanism: Design, modeling, control, and
  performance evaluation,'' \emph{The International Journal of Robotics
  Research}, vol.~36, no.~4, pp. 414--435, 2017.

\bibitem{LvGregg2018TCST}
G.~{Lv} and R.~D. {Gregg}, ``Underactuated potential energy shaping with
  contact constraints: Application to a powered knee-ankle orthosis,''
  \emph{IEEE Transactions on Control Systems Technology}, vol.~26, no.~1, pp.
  181--193, 2018.

\bibitem{LeeLeeKimHanShinHan2014Mechatronics}
H.-D. Lee, B.-K. Lee, W.-S. Kim, J.-S. Han, K.-S. Shin, and C.-S. Han,
  ``Human--robot cooperation control based on a dynamic model of an upper limb
  exoskeleton for human power amplification,'' \emph{Mechatronics}, vol.~24,
  no.~2, pp. 168--176, 2014.

\bibitem{BuergerHogan2007TRO}
S.~P. Buerger and N.~Hogan, ``Complementary stability and loop shaping for
  improved human--robot interaction,'' \emph{IEEE Transactions on Robotics},
  vol.~23, no.~2, pp. 232--244, 2007.

\bibitem{HeThomasPaineSentis2019ACC}
B.~He, G.~C. Thomas, N.~Paine, and L.~Sentis, ``Modeling and loop shaping of
  single-joint amplification exoskeleton with contact sensing and series
  elastic actuation,'' in \emph{2019 American Control Conference (ACC)}.\hskip
  1em plus 0.5em minus 0.4em\relax IEEE, 2019, pp. 4580--4587.

\bibitem{HuangCappelThomasHeSentis2020ACC}
H.~Huang, H.~F. Cappel, G.~C. Thomas, B.~He, and L.~Sentis, ``Adaptive
  compliance shaping with human impedance estimation,'' in \emph{2020 American
  Control Conference (ACC)}.\hskip 1em plus 0.5em minus 0.4em\relax IEEE, 2020,
  pp. 5131--5138.

\bibitem{RouseHargrovePerreaultKuiken2014TNSRE}
E.~J. Rouse, L.~J. Hargrove, E.~J. Perreault, and T.~A. Kuiken, ``Estimation of
  human ankle impedance during the stance phase of walking,'' \emph{IEEE
  Transactions on Neural Systems and Rehabilitation Engineering}, vol.~22,
  no.~4, pp. 870--878, 2014.

\bibitem{LeeHogan2015TNSRE}
H.~Lee and N.~Hogan, ``Time-varying ankle mechanical impedance during human
  locomotion,'' \emph{IEEE Transactions on Neural Systems and Rehabilitation
  Engineering}, vol.~23, no.~5, pp. 755--764, 2015.

\bibitem{LeeRouseKrebs2016JTEHM}
H.~Lee, E.~J. Rouse, and H.~I. Krebs, ``Summary of human ankle mechanical
  impedance during walking,'' \emph{IEEE journal of translational engineering
  in health and medicine}, vol.~4, pp. 1--7, 2016.

\bibitem{ShorterRouse2018TNSRE}
A.~L. {Shorter} and E.~J. {Rouse}, ``Mechanical impedance of the ankle during
  the terminal stance phase of walking,'' \emph{IEEE Transactions on Neural
  Systems and Rehabilitation Engineering}, vol.~26, no.~1, pp. 135--143, 2018.

\bibitem{ShorterRouse2019TBME}
------, ``Ankle mechanical impedance during the stance phase of running,''
  \emph{IEEE Transactions on Biomedical Engineering}, pp. 1--1, 2019.

\bibitem{BennettHollerbachXuHunter1992EBR}
D.~Bennett, J.~Hollerbach, Y.~Xu, and I.~Hunter, ``Time-varying stiffness of
  human elbow joint during cyclic voluntary movement,'' \emph{Experimental
  Brain Research}, vol.~88, no.~2, pp. 433--442, 1992.

\bibitem{RouseGreggHargroveSensinger2012TBE}
E.~J. Rouse, R.~D. Gregg, L.~J. Hargrove, and J.~W. Sensinger, ``The difference
  between stiffness and quasi-stiffness in the context of biomechanical
  modeling,'' \emph{IEEE Transactions on Biomedical Engineering}, vol.~60,
  no.~2, pp. 562--568, 2012.

\bibitem{AgarwalGottlieb1977JBE}
G.~Agarwal and C.~Gottlieb, ``Compliance of the human ankle joint,''
  \emph{Journal of Biomechanical Engineering}, vol.~99, no.~3, pp. 166--170,
  1977.

\bibitem{GottliebAgarwal1978JB}
G.~L. Gottlieb and G.~C. Agarwal, ``Dependence of human ankle compliance on
  joint angle,'' \emph{Journal of Biomechanics}, vol.~11, no.~4, pp. 177--181,
  1978.

\bibitem{ZahalakHeyman1979JBE}
G.~Zahalak and S.~Heyman, ``A quantitative evaluation of the frequency-response
  characteristics of active human skeletal muscle in vivo,'' \emph{Journal of
  Biomechanical Engineering}, vol. 101, pp. 28--37, 1979.

\bibitem{CannonZahalak1982JB}
S.~C. Cannon and G.~I. Zahalak, ``The mechanical behavior of active human
  skeletal muscle in small oscillations,'' \emph{Journal of Biomechanics},
  vol.~15, no.~2, pp. 111--121, 1982.

\bibitem{HunterKearney1982JB}
I.~Hunter and R.~Kearney, ``Dynamics of human ankle stiffness: Variation with
  mean ankle torque,'' \emph{Journal of Biomechanics}, vol.~15, no.~10, pp. 747
  -- 752, 1982.

\bibitem{BeckerMote1990JBE}
J.~Becker and C.~Mote, ``Identification of a frequency response model of joint
  rotation,'' \emph{Journal of Biomechanical Engineering}, vol. 112, no.~1, pp.
  1--8, 1990.

\bibitem{WeissHunterKearney1988JB}
P.~Weiss, I.~Hunter, and R.~Kearney, ``Human ankle joint stiffness over the
  full range of muscle activation levels,'' \emph{Journal of Biomechanics},
  vol.~21, no.~7, pp. 539--544, 1988.

\bibitem{SobhanitehraniJalaleddiniKearney2017TNSRE}
E.~{Sobhani Tehrani}, K.~{Jalaleddini}, and R.~E. {Kearney}, ``Ankle joint
  intrinsic dynamics is more complex than a mass-spring-damper model,''
  \emph{IEEE Transactions on Neural Systems and Rehabilitation Engineering},
  vol.~25, no.~9, pp. 1568--1580, 2017.

\bibitem{PerreaultKirschCrago2004EBR}
E.~J. Perreault, R.~F. Kirsch, and P.~E. Crago, ``Multijoint dynamics and
  postural stability of the human arm,'' \emph{Experimental Brain Research},
  vol. 157, no.~4, pp. 507--517, 2004.

\bibitem{MilnerCloutier1993EBR}
T.~E. Milner and C.~Cloutier, ``Compensation for mechanically unstable loading
  in voluntary wrist movement,'' \emph{Experimental Brain Research}, vol.~94,
  no.~3, pp. 522--532, 1993.

\bibitem{BishopJohnson2011book}
R.~E.~D. Bishop and D.~C. Johnson, \emph{The Mechanics of Vibration}.\hskip 1em
  plus 0.5em minus 0.4em\relax Cambridge University Press, 1960.

\bibitem{HeHuangThomasSentis2019IROS}
B.~{He}, H.~{Huang}, G.~C. {Thomas}, and L.~{Sentis}, ``Complex stiffness model
  of physical human-robot interaction: Implications for control of performance
  augmentation exoskeletons,'' in \emph{2019 IEEE/RSJ International Conference
  on Intelligent Robots and Systems (IROS)}.\hskip 1em plus 0.5em minus
  0.4em\relax IEEE, 2019, pp. 6748--6755.

\bibitem{PaineOhSentis2014TMech}
N.~Paine, S.~Oh, and L.~Sentis, ``Design and control considerations for
  high-performance series elastic actuators,'' \emph{IEEE/ASME Transactions on
  Mechatronics}, vol.~19, no.~3, pp. 1080--1091, 2014.

\bibitem{PopeCrowninshieldMillerJohnson1976JB}
M.~Pope, R.~Crowninshield, R.~Miller, and R.~Johnson, ``The static and dynamic
  behavior of the human knee in vivo,'' \emph{Journal of biomechanics}, vol.~9,
  no.~7, pp. 449--452, 1976.

\bibitem{PoppeleBowman1970JN}
R.~Poppele and R.~Bowman, ``Quantitative description of linear behavior of
  mammalian muscle spindles.'' \emph{journal of Neurophysiology}, vol.~33,
  no.~1, pp. 59--72, 1970.

\bibitem{WesterhofNoordergraaf1970JB}
N.~Westerhof and A.~Noordergraaf, ``Arterial viscoelasticity: a generalized
  model: effect on input impedance and wave travel in the systematic tree,''
  \emph{Journal of Biomechanics}, vol.~3, no.~3, pp. 357--379, 1970.

\bibitem{InaudiKelly1995JEM}
J.~A. Inaudi and J.~M. Kelly, ``Linear hysteretic damping and the hilbert
  transform,'' \emph{Journal of Engineering Mechanics}, vol. 121, no.~5, pp.
  626--632, 1995.

\end{thebibliography}

\end{document}